\documentclass[letterpaper, 10 pt, conference]{ieeeconf}  

\IEEEoverridecommandlockouts                              

\usepackage{times}

\usepackage[dvipsnames]{xcolor}
\usepackage{moreverb,url}
\usepackage{amsmath,amsfonts,amssymb}
\usepackage{tikz,tkz-euclide}
\usepackage{balance}
\usetikzlibrary{decorations,calligraphy}
\usetikzlibrary{calc}
\tikzset{>=latex}

\usepackage[nolist,nohyperlinks]{acronym}

\def\m{m}

\def\reals{\mathbb{R}}

\renewcommand\time[1]{{t_{#1}}}

\newcommand\rot[2]{{\mathbf{R}_{#1}^{#2}}}

\newcommand\pos[2]{{\mathbf{p}_{#1}^{#2}}}

\newcommand\vel[2]{{\mathbf{v}_{#1}^{#2}}}

\def\world{{W}}

\newcommand\angvel[1]{{\omega_{#1}}}

\def\coordinate{{\mathbf{e}}}

\newcommand\encoder[1]{{c_{#1}}}
\def\wheelradius{{r}}
\newcommand\orientation[2]{{\theta_{#1}^{#2}}}
\newcommand\gtangvel[1]{{\omega_{GT}^{#1}}}
\newcommand\veltilde[2]{{\tilde{\mathbf{v}}_{#1}^{#2}}}

\begin{acronym}[AAAAAAAAA]
    \acro{1d}[1D]{One-Dimensional}
    \acro{2d}[2D]{Two-Dimensional}
    \acro{2fast}[2Fast-2Lamaa]{Fast Field-based Agent-Subtracted Tightly-coupled Lidar Localisation And Mapping with Accelerometer and Angular-rate}
    \acro{3d}[3D]{Three-Dimensional}
    \acro{adas}[ADAS]{Advanced Driver-Assistance System}
    \acro{awd}[AWD]{All-Wheel-Drive}
    \acro{cas}[CAS]{Centre for Autonomous Systems}
    \acro{cnn}[CNN]{Convolutional Neural Network}
    \acro{cpu}[CPU]{Central Processing Unit}
    \acro{doals}[DOALS]{Urban Dynamic Objects LiDAR}
    \acro{dof}[DoF]{Degree-of-Freedom}
    \acro{dro}[DRO]{Direct Radar Odometry}
    \acro{dvs}[DVS]{Dynamic Vision Sensor}
    \acrodefplural{dvs}[DVS's]{Dynamic Vision Sensors}
    \acro{ekf}[EKF]{Extended Kalman filter}
    \acro{fifo}[FIFO]{First In, First Out}
    \acro{fmcw}[FMCW]{Frequency-Modulated Continuous-Wave}
    \acro{fov}[FoV]{Field-of-View}
    \acro{fwd}[FWD]{Front-Wheel-Drive}
    \acro{gnss}[GNSS]{Global Navigation Satellite System}
    \acrodefplural{gnss}[GNSS's]{Global Navigation Satellite Systems}
    \acro{gp}[GP]{Gaussian Process}
    \acrodefplural{gp}[GPs]{Gaussian Processes}
    \acro{gpm}[GPM]{Gaussian Preintegrated Measurement}
    \acro{ugpm}[UGPM]{Unified Gaussian Preintegrated Measurement}
    \acro{gps}[GPS]{Global Position System}
    \acrodefplural{gps}[GPS's]{Global Position Systems}
    \acro{gpu}[GPU]{Graphic Processing Unit}
    \acro{hdr}[HDR]{High Dynamic Range}
    \acro{hri}[HRI]{Human-Robot Interactions}
    \acro{icp}[ICP]{Iterative Closest Point}
    \acro{idol}[IDOL]{IMU-DVS Odometry using Lines}
    \acro{iou}[IoU]{Intersection over Union}
    \acro{imu}[IMU]{Inertial Measurement Unit}
    \acro{in2laama}[IN2LAAMA]{INertial Lidar Localisation Autocalibration And MApping}
    \acro{kf}[KF]{Kalman Filter}
    \acro{kl}[KL]{Kullback–Leibler}
    \acro{lidar}[LiDAR]{Light Detection And Ranging Sensor}
    \acro{lpm}[LPM]{Linear Preintegrated Measurement}
    \acro{map}[MAP]{Maximum A Posteriori}
    \acro{mle}[MLE]{Maximum Likelihood Estimation}
    \acro{mmwave}[mmWave]{millimeter wave}
    \acro{ndt}[NDT]{Normal Distribution Transform}
    \acro{og}[OG]{Odometer-Gyroscope}
    \acro{pca}[PCA]{Principal Component Analysis}
    \acro{pm}[PM]{Preintegrated Measurement}
    \acro{rcnn}[R-CNN]{Region-based Convolutional Neural Network}
    \acro{rgb}[RGB]{color}
    \acro{rgbd}[RGBD]{color-depth}
    \acro{rms}[RMS]{Root Mean Squared}
    \acro{rmse}[RMSE]{Root Mean Squared Error}
    \acro{ros}[ROS]{Robot Operating System}
    \acro{rpe}[RPE]{Relative Position Error}
    \acro{rrbt}[RRBT]{Rapidly Exploring Random Belief Trees}
    \acro{sde}[SDE]{Stochastic Differential Equation}
    \acro{SE3}[SE(3)]{Special Euclidean group in three dimensions}
    \acro{slam}[SLAM]{Simultaneous Localisation And Mapping}
    \acro{so3}[$\mathfrak{so}$(3)]{Lie algebra of special orthonormal group in three dimensions}
    \acro{SO3}[SO(3)]{Special Orthonormal rotation group in three dimensions}
    \acro{sota}[SotA]{state-of-the-art}
    \acro{upm}[UPM]{Upsampled-Preintegrated-Measurement}
    \acro{uts}[UTS]{University of Technology, Sydney}
    \acro{vi}[VI]{Visual-Inertial}
    \acro{vio}[VIO]{Visual-Inertial Odometry}
    \acro{vo}[VO]{Visual Odometry}
\end{acronym}
\usepackage[bookmarks=true]{hyperref}
\usepackage{multicol}
\usepackage{booktabs}
\usepackage[weather]{ifsym}
\usepackage{siunitx}
\usepackage{tabularx}
\newcolumntype{Y}{>{\centering\arraybackslash}X}

\usepackage{fancyhdr}
\fancypagestyle{withfooter}{
  
  \fancyhead[L]{}
  \fancyhead[R]{}
  \fancyfoot[C]{\footnotesize Presented at the 2025 IEEE ICRA Workshop on Field Robotics}
}

\title{\LARGE \bf
Do We Still Need to Work on Odometry for Autonomous Driving?
}

\author{Cedric Le Gentil, Daniil Lisus, and Timothy D. Barfoot 
\thanks{This paper was partially supported by an Ontario Research Fund - Research Excellence grant.}
\thanks{All authors are with the Autonomous Space Robotics Lab at the University of Toronto Institute for Aerospace Studies, Toronto, ON, Canada.
        {\tt\small cedric.legentil@robotics.utias.utoronto.ca}}%
}

\begin{document}

\maketitle
\thispagestyle{withfooter}
\pagestyle{withfooter}

\begin{abstract}

Over the past decades, a tremendous amount of work has addressed the topic of ego-motion estimation of moving platforms based on various proprioceptive and exteroceptive sensors.
At the cost of ever-increasing computational load and sensor complexity, odometry algorithms have reached impressive levels of accuracy with minimal drift in various conditions.
In this paper, we question the need for more research on odometry for autonomous driving by assessing the accuracy of one of the simplest algorithms: the direct integration of wheel encoder data and yaw rate measurements from a gyroscope.
We denote this algorithm as Odometer-Gyroscope (OG) odometry.
This work shows that OG odometry can outperform current state-of-the-art radar-inertial SE(2) odometry for a fraction of the computational cost in most scenarios.
For example, the OG odometry is on top of the Boreas leaderboard with a relative translation error of 0.20\%, while the second-best method displays an error of 0.26\%.
Lidar-inertial approaches can provide more accurate estimates, but the computational load is three orders of magnitude higher than the OG odometry. 
To further the analysis, we have pushed the limits of the OG odometry by purposely violating its fundamental no-slip assumption using data collected during a heavy snowstorm with different driving behaviours.
Our conclusion shows that a significant amount of slippage is required to result in non-satisfactory pose estimates from the OG odometry.

\end{abstract}

\section{Introduction}

Spatial awareness is a crucial component of any autonomous system.
Over the past two decades, the robotics community put a colossal amount of work into addressing the problem of localization in unknown environments via odometry and \ac{slam}~\cite{cadena2016past}.
While many techniques are generic in terms of applications (UAVs, off-road robots, handheld devices, etc.), a large number of them have been tailored to the problem of autonomous driving and rely on relatively expensive sensor packages.
Since the release of the KITTI dataset~\cite{geiger2012kitti}, a significant drive of the odometry and \ac{slam} research has been the improvement of trajectory accuracy in various benchmarks using lidar, visual, and, more recently, radar data as the main sensing modalities~\cite{venon2022millimeter,nader2024survey}. 
With this work, we want to take a step back and reflect on the abilities of state-of-the-art methods in the context of autonomous driving with respect to approaches that are more efficient in terms of cost and computation.
Focusing on SE(2) odometry, we consider one of the simplest sensor suites and algorithms by combining a wheel encoder with a yaw gyroscope and performing naive integration of the sensory data.
Not claiming any novelty here, we will refer to this approach as the \ac{og} odometry.

A clear limitation of encoder-based odometry is the sensitivity to the non-slip assumption.
Any slippage between the wheel and the road introduces an unrecoverable error in the odometry estimate.
This partly motivates the use of exteroceptive sensors, which are inherently agnostic to slippage.
Cameras and lidars are the most popular modalities used for autonomous driving perception.
However, these can be affected by adverse conditions such as heavy rain or snow with various levels of occlusion, thus limiting their effectiveness and robustness in extreme scenarios.
Recently, radars have regained popularity within the robotics community~\cite{harlow2024newwave}.
As millimetre waves are not affected by precipitation or dust, they are seen as the sensor of choice for all-weather autonomous driving.
Accordingly, we chose to benchmark the \ac{og} odometry mainly with \ac{sota} radar-based odometry frameworks.
Our analysis includes one lidar-inertial baseline.

\begin{figure}
    \centering
    \def\vdist{2.6cm}
    \def\hdist{0.05cm}
    \def\legenddist{0.00cm}
    \def\legendtextsize{\small}
    \begin{tikzpicture}
        \node[inner sep=0, outer sep=0](photo){\includegraphics[width=0.99\columnwidth]{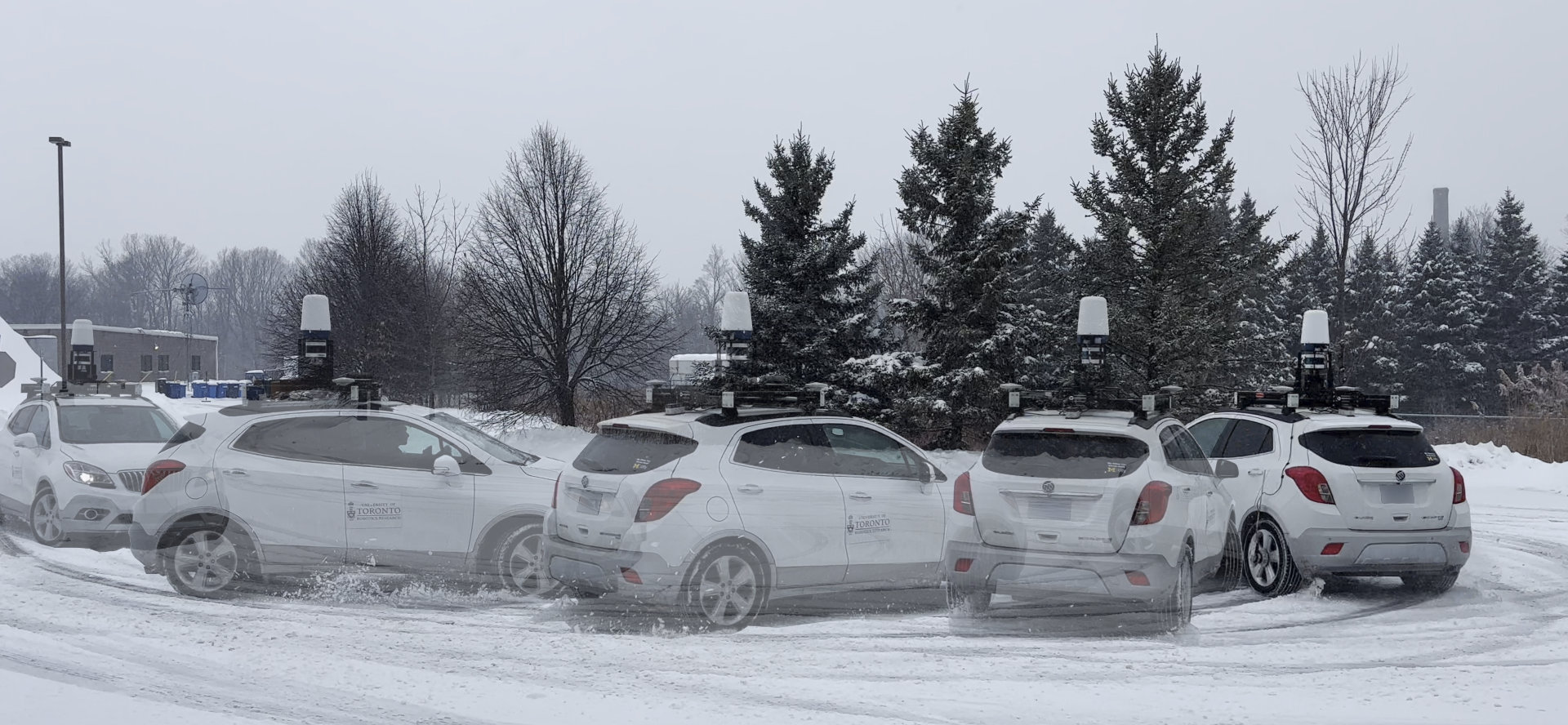}};
        \coordinate[below=\vdist of photo] (mid);
        \node[inner sep=0, outer sep=0, left=\hdist of mid](noslip){\includegraphics[width=0.49\columnwidth]{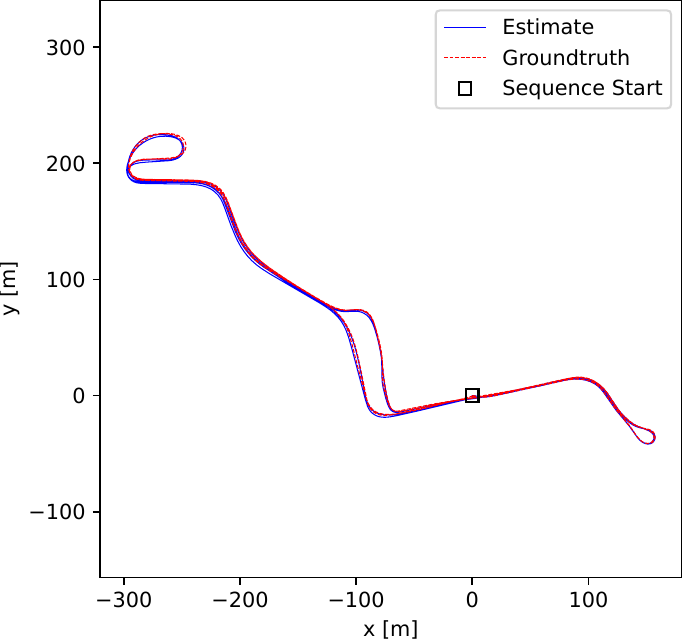}};
        \node[inner sep=0, outer sep=0, right=\hdist of mid](slip){\includegraphics[width=0.49\columnwidth]{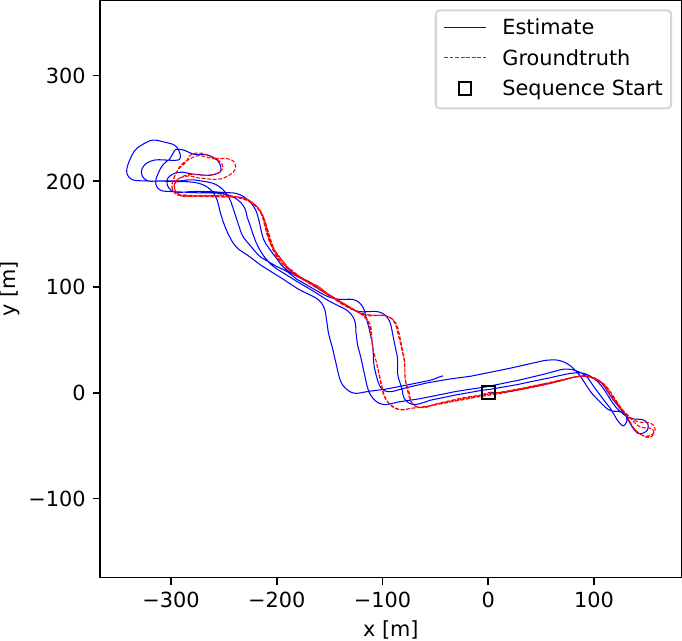}};

        \node[below=\legenddist of photo]{\legendtextsize (a) Novel data with challenging trajectories};
        \node[below=\legenddist of noslip]{\legendtextsize (b) No-slip trajectory};
        \node[below=\legenddist of slip]{\legendtextsize (c) High-slip trajectory};

    \end{tikzpicture}
    \caption{This paper benchmarks a simple wheel-gyroscope odometry method, denoted OG, against state-of-the-art (SotA) radar-based pipelines. When considering nominal scenarios, the OG odometry shows SotA performance (b). To push the limits of the OG odometry and further our analysis, we have collected data during a snowstorm and purposely introduced slippage in the vehicle's trajectory (a)(c).}
    \label{fig:teaser}
\end{figure}

This work highlights that wheel-based odometry is not `a thing of the past' or `unreliable'.
Through the use of public data across a wide range of weather conditions and data collected specifically for this analysis, our experiments reveal that the \ac{og} odometry can outperform \ac{sota} radar frameworks.
The novel data pushes the limits of the \ac{og} odometry by purposely violating the no-slip assumption, as shown in Fig.~\ref{fig:teaser}.
In addition to providing competitive accuracy, \ac{og} odometry presents other advantages, such as an extremely lightweight computational load and robustness to dynamic environments.

After a brief review of the relevant lidar and radar-based odometry literature in Section~\ref{sec:related}, we describe the \ac{og} odometry in Section~\ref{sec:og}.
Section~\ref{sec:experiments} details our experimental setup and shows the results obtained with \ac{og} and \ac{sota} odometry frameworks.
In Section~\ref{sec:discussion}, we open the discussion around the relevance of odometry research for autonomous driving before concluding in Section~\ref{sec:conclusion}.

\section{Related work}
\label{sec:related}

Thanks to technology improvements and a decreasing price, lidars became ubiquitous in state estimation for robotics with numerous ready-to-use open-source frameworks~\cite{lee2024lidar}.
A key component of dealing with lidar data is accounting for motion distortion.
While approaches such as~\cite{ye2019tightly, shan2020liosam,chen2023dlio} perform motion correction as a pre-processing step based on the previous state estimate and inertial data, methods based on some kind of continuous representation provide a more elegant solution to lidar-based state estimation.
For example,~\cite{bosse2012zebedee} uses piece-wise linear functions to represent the system's pose.
The B-Spline representation from~\cite{Furgale2015} is used in~\cite{droeschel2018efficient}.
The continuous-time ICP presented in~\cite{are_we_ready_for} is built on the \ac{gp} trajectory representation from~\cite{Anderson2015FullSTEAM} with a white-noise-on-acceleration motion prior.
Alternatively, using continuous \ac{gp}-based \ac{imu} preintegration~\cite{legentil2020gpm}, \cite{legentil2021in2laama} introduces a hybrid continuous-discrete approach to state estimation.
More recently,~\cite{legentil2024truly} proposed to parameterize the lidar trajectory locally solely using the \ac{imu} biases based on the continuous preintegration from~\cite{legentil2023latent}.

While radar-based sensing for mobile robotics has recently regained popularity~\cite{harlow2024newwave}, \ac{fmcw} radars are not new to the field.
In the early days of \ac{slam}, metallic reflectors were spread throughout the environment and detected in the radar data to be used as landmarks for state estimation~\cite{dissanayake2001slam}.
As the sensors' precision and the available computational power drastically increased, newer ego-motion estimation algorithms reached levels of accuracy that allow safe and robust autonomous operation of mobile robots in various environments~\cite{are_we_ready_for,qiao2025radar}.
The appeal of radar data is partly due to its robustness to weather conditions.
Unlike lidars and cameras that require clear line-of-sight, radar can perceive the environment through dust, fog, and heavy precipitation (rain or snow).
As electromagnetic waves can penetrate and pass through some materials, the raw radar data from spinning radars consists of polar images in which every pixel's intensity corresponds to the wave reflection at a given range and azimuth.

A standard radar-based state estimation strategy is to extract features from the radar intensity data before performing point cloud registration.
In early odometry works such as~\cite{callmer2011radarslamusingvisualfeatures}, SIFT features~\cite{lowe2004sift} were extracted from the radar images in their Cartesian form.
More recent approaches, such as CFEAR~\cite{adolfsson2023cfear}, use the radar polar images to obtain point clouds directly.
The authors of~\cite{pprestonkrebs2024thefinerpoints} conduct a detailed analysis of the impact of feature selection on odometry accuracy.
In order to perform scan-to-scan or scan-to-map registration, feature-based methods need to perform data association.
Either with point-to-point associations~\cite{2021_Burnett, hero_paper,burnett2024continuous} or point-to-line~\cite{adolfsson2023cfear}, feature-based odometry is prone to outlier associations, especially in the presence of dynamic objects in the scene, and generally requires the use of robust loss functions.
In~\cite{Kubelka2024} and \cite{lisus2025doppler}, the Doppler effect is used to help or replace association-based registration.
Direct methods such as~\cite{masking_by_moving,Checchin2009,park2020pharao,Rennie_Williams_Newman_DeMartini_2023} partly alleviate the dynamic object issue by using all the data in the radar scans.
Unfortunately, these methods discretize the state variables for the sake of computation load, thus impeding the state estimation accuracy.
More recently, DRO~\cite{legenti2025dro} introduces a direct radar odometry method that considers the sensor's continuous-time trajectory in a scan-to-local-map registration approach.
While direct methods are less affected by moving objects, some scenarios are still challenging when the majority of the radar intensity data corresponds to moving vehicles.
Using solely proprioceptive sensors, the \ac{og} odometry does not necessitate any form of data association and is completely agnostic to dynamicity in the environment.

Works such as~\cite{hartley2020contact,niu2021wheel,wu2023wheel} show how the knowledge of the system's kinematic and an \ac{imu} can be used for accurate odometry without the need for exteroception.
Unfortunately, despite the profusion of automotive-oriented datasets that span a wide range of environments and weather conditions, for example~\cite{maddern2017oxford,jeong2019complex,kim2020mulran,gadd2024oord,kim2025hercules}, few provide data from both a wheel encoder and an \ac{imu}. The exception to this is the Boreas dataset~\cite{burnett2023boreas}, which we use for many of the experiments in this paper.
To the best of our knowledge, no publicly available large-scale automotive dataset contains challenging vehicle trajectories that violate the standard no-slip assumption.
With this work, we highlight that given a simple encoder and a gyroscope, odometry can be performed at the level of~\ac{sota} exteroceptive methods for virtually no computational load.
Thus, we claim that the robotics community needs more challenging datasets to focus on actually challenging state estimation problems that cannot be solved with one of the simplest sensor suites and algorithms.

\section{Odometer-Gyroscope odometry}
\label{sec:og}

\subsection{Odometry}

Let us consider a vehicle equipped with a simple encoder mounted on a non-steering wheel (e.g., on the back wheel of a standard car) and a heading gyroscope.
Without loss of generality, let us consider sensors that are synchronized and provide measurements at the same time.
The encoder and gyroscope, respectively, provide the cumulative counter value $\encoder{i}$ and the yaw angular velocity $\angvel{i}$ at time $\time{i}$.
The wheel radius is denoted $\wheelradius$, and $N$ is the number of encoder ticks per revolution.
The \ac{og} odometry aims at estimating the wheel's center 2D pose through time with respect to a world frame $\world$.
We denote the wheel's position $\pos{\world}{\time{i}} \in \reals^2$, and its orientation $\orientation{\world}{\time{i}} \in \reals$.

The distance $d_i$ traveled by the wheel center between $\time{i+1}$ and $\time{i}$ is given by
\begin{equation}
    d_i = \frac{2\pi\wheelradius}{N}(\encoder{i+1} - \encoder{i}).
    \label{eq:dist}
\end{equation}
Starting with $\pos{\world}{\time{0}} = \mathbf{0}$ and $\orientation{\world}{\time{0}} = 0$, the odometry process is defined as
\begin{align}
    \pos{\world}{\time{i+1}} &= \begin{bmatrix} \cos(\orientation{\world}{\time{i}}) & -\sin(\orientation{\world}{\time{i}}) \\ \sin(\orientation{\world}{\time{i}}) & \cos(\orientation{\world}{\time{i}}) \end{bmatrix} \Delta\pos{}{},
    \\
    \orientation{\world}{\time{i+1}} &= \orientation{\world}{\time{i}} + \Delta\orientation{}{},
\end{align}
with
\begin{equation}
    \Delta\pos{}{} = \begin{cases} \begin{bmatrix}
        d_i \\ 0 \end{bmatrix} & \text{if}\ \Delta\orientation{}{} \approx 0 
        \\
        \\
        \frac{d_i}{\Delta\orientation{}{}}\begin{bmatrix}
        \sin(\Delta\orientation{}{}) \\ 1-\cos(\Delta\orientation{}{}) \end{bmatrix} & \text{otherwise} \end{cases},
\end{equation}
and
\begin{equation}
    \Delta\orientation{}{} = \frac{(\angvel{i} + \angvel{i+1})(\time{i+1} - \time{i})}{2}.
\end{equation}

\subsection{Gyroscope bias}

Unfortunately, in real-world systems, the gyroscope measurements are affected by a slow-varying additive bias $b$.
Thus, we model a gyroscope measurement at $t_i$ as $\tilde{\angvel{i}} = \angvel{i} + b_i$.
To limit the impact of the bias on the \ac{og} odometry, we adopt a simple yet effective bias estimation strategy based on the heuristic that when the vehicle is static, the vehicle's angular velocity is null.
Accordingly, we initialize the bias by averaging the raw gyroscope measurements using 2 seconds of data when the encoder indicates no motion ($\encoder{i} = \encoder{i+1}$).
To account for the slow-varying nature of the bias, we update that estimate with a first-order low-pass filter.

\subsection{Calibration}

The accuracy of the \ac{og} odometry is linearly tied to the accuracy of the wheel radius $\wheelradius$ used in \eqref{eq:dist}.
We propose to estimate this value using ground-truth body-centric linear and angular velocity information denoted $\vel{GT}{\time{i}}$ and $\gtangvel{\time{i}}$ (measured with an independent GNSS/INS solution).
From the wheel encoder, we obtain the body-centric velocity information as $\veltilde{w}{\time{i}} = \begin{bmatrix}\frac{d_i}{\time{i+1}-\time{i}} & 0\end{bmatrix}^\top$.
The ground-truth and encoder velocity are linked as
\begin{equation}
    \veltilde{GT}{\time{i}} = \rot{GT}{w}\veltilde{w}{\time{i}} + \begin{bmatrix} 0 & -\gtangvel{\time{i}} \\ \gtangvel{\time{i}} & 0 \end{bmatrix} \pos{GT}{w},
\end{equation}
where $\pos{GT}{w}$ and $\rot{GT}{w}$ are the position and orientation of the wheel center in the GNSS/INS reference frame and represent the extrinsic calibration parameters between the wheel and the ground-truth solution.
While one can have prior knowledge about $\pos{GT}{w}$ and $\rot{GT}{w}$, we choose to estimate these extrinsic parameters alongside the wheel radius through a non-linear least-squares optimization:
\begin{equation}
    \wheelradius^*, \pos{GT}{w}^*, \rot{GT}{w}^* = \underset{\wheelradius, \pos{GT}{w}, \rot{GT}{w}}{\text{argmin}}\quad \sum \Vert\vel{GT}{\time{i}} - \veltilde{GT}{\time{i}}\Vert^2.
\end{equation}
Note that the knowledge of $\pos{GT}{w}$ and $\rot{GT}{w}$ is also required for the final evaluation of the \ac{og} odometry as the ground-truth trajectory corresponds to the GNSS/INS pose.

\section{Experiments}
\label{sec:experiments}

\subsection{Nominal conditions: Boreas dataset}
\label{sec:boreas_exp}

\subsubsection{Dataset}

To compare the \ac{og} odometry with \ac{sota} radar-based odometry frameworks, we leverage the Boreas dataset~\cite{burnett2023boreas} and its leaderboard.
The data consists of 44 sequences collected repeatedly over one year with a 3D lidar, a 2D spinning radar, a front-facing camera, and an RTK-GNSS/INS/encoder solution for ground-truth generation.
The ground-truth trajectory is not provided for 13 of the sequences, as these are used to benchmark the different methods on the leaderboard.
The 13 sequences span a wide range of weather conditions from bright sunny days to heavy snowfalls.
While some sequences have been recorded with high-snow coverage, no significant slippage occurs in this dataset.

\subsubsection{Baselines}

The Boreas dataset has been used for a workshop competition in 2024~\cite{workshop2024radar}, and the results have been integrated into the leaderboard.
While the rules allowed for the use of encoder data, none of the methods currently on the leaderboard leverage this piece of information.
We have selected the four best-performing pipelines that are DRO-G, CFEAR++~\cite{li2024cfearpp}, CFEAR~\cite{adolfsson2023cfear}, and STEAM-RIO++~\cite{burnett2024continuous}.
All these methods, except for DRO-G, rely on point-cloud extraction from the raw radar data and perform ICP-based scan registration in a sliding-window manner.
DRO-G directly uses the radar intensity returns with a scan-to-local-map registration.
All methods except CFEAR use information from the INS' \ac{imu}.
Accordingly, it is important to note that inertial-aided methods might be positively biased as the INS' \ac{imu} is used in the ground-truth generation process.

\subsubsection{Results}

\begin{table}[]
    \centering
    \caption{Boreas dataset SE(2) odometry leaderboard.}
    \vspace{-0.3cm}
    \begin{tabularx}{\columnwidth}{lYY}
        \toprule
        \textbf{Method} & \textbf{Trans. err. [\%]} & \textbf{Rot. err. [${}^\circ/100\mathrm{m}$]} 
        \\
        \midrule
        OG odometry & \textbf{0.20} & \textbf{0.03}
        \\
        DRO-G & 0.26 & 0.05
        \\ 
        CFEAR++ \cite{li2024cfearpp} & 0.51 & 0.14
        \\
        CFEAR \cite{adolfsson2023cfear} & 0.61 & 0.21
        \\
        STEAM-RIO++ \cite{burnett2024continuous} & 0.62 & 0.18
        \\
        \bottomrule
    \end{tabularx}
    \label{tab:boreas_baseline}
\end{table}

\begin{table}[]
    \centering
    \caption{Per-sequence errors of the OG odometry on the Boeras dataset leaderboard.}
    \vspace{-0.3cm}
    \label{tab:boreas_detailed}
    \begin{tabular}{lccc}
        \toprule
        \multicolumn{1}{c}{\textbf{\shortstack{Sequence\\ID}}} & \multicolumn{1}{c}{\textbf{\shortstack{Weather\\conditions}}} & \multicolumn{1}{c}{\textbf{\shortstack{Trans.\\err. [\%]}}} & \multicolumn{1}{c}{\textbf{\shortstack{Rot. err. \\ $[{}^\circ/100\mathrm{m}]$}}}
        \\
        \midrule
        2020-12-04-14-00 & \FilledCloud\Snow & 0.24 & 0.05
        \\ 
        2021-01-26-10-59 & \FilledSnowCloud\Snow\Snow & 0.37 & 0.01
        \\ 
        2021-02-09-12-55 & \FilledSunCloud\Snow & 0.20 & 0.02
        \\ 
        2021-03-09-14-23 & \Sun & 0.14 & 0.02
        \\ 
        2021-04-22-15-00 & \FilledSnowCloud & 0.12 & 0.02
        \\ 
        2021-06-29-18-53 & \FilledRainCloud & 0.11 & 0.02
        \\ 
        2021-06-29-20-43 & \Cloud\HalfSun & 0.13 & 0.02
        \\ 
        2021-09-08-21-00 & \NoSun & 0.14 & 0.02
        \\ 
        2021-09-09-15-28 & \SunCloud & 0.34 & 0.06
        \\ 
        2021-10-05-15-35 & \FilledCloud & 0.14 & 0.03
        \\ 
        2021-10-26-12-35 & \FilledRainCloud & 0.13 & 0.02
        \\ 
        2021-11-06-18-55 & \NoSun & 0.26 & 0.05
        \\ 
        2021-11-28-09-18 & \FilledSnowCloud\Snow\Snow & 0.32 & 0.04
        \\
        \bottomrule
        \multicolumn{4}{c}{\FilledCloud: Overcast, \Snow: Snow coverage, \Snow\Snow: High snow coverage}
        \\
        \multicolumn{4}{c}{ \FilledSnowCloud: Snowing, \Sun: Sun, \FilledRainCloud: Rain, \HalfSun: Dusk, \NoSun: Night}
    \end{tabular}
\end{table}

Table~\ref{tab:boreas_baseline} presents the results obtained on the Boreas leaderboard with the standard KITTI odometry metrics (relative translation error [\%] and relative rotation error [$^\circ/\si{100m}$]) as detailed in~\cite{burnett2023boreas}.
The results obtained with the \ac{og} odometry use wheel radius and extrinsic parameters computed based on 14 of the sequences that possess publicly available ground-truth.
The \ac{og} odometry outperforms any other method.
To be completely fair with the other methods, the \ac{og} odometry results could benefit from the fact that the wheel encoder is also weakly used in the generation of the ground-truth trajectory, thus slightly biasing the results in a positive manner.
However, one can see in Table~\ref{tab:boreas_detailed} that the relative translation error seems to be higher for high-snow coverage sequences.
This correlates with the increased likelihood of slippage when driving over snow-covered roads.
Associated with the fact that other methods use the INS's \ac{imu}, we believe that regardless of the potential bias, the \ac{og} odometry displays \ac{sota} performance.
We test these assumptions further with our next set of experiments, which have been carefully designed to avoid any potential ground-truth biases in the evaluation.

\subsection{Challenging conditions}

\subsubsection{Dataset}

To further the analysis of the \ac{og} odometry, we have collected a dataset that violates the no-slip assumption with the platform used in the Boreas dataset~\cite{burnett2023boreas}.
The only differences are changes in the firmware of the Navtech radar that allow for Doppler-based velocity estimation, the addition of a stand-alone 6-DoF \ac{imu} (Silicon Sensing DMU41) that is not coupled to the GNSS/INS solution, and the fact that the ground-truth trajectory is generated without the wheel encoder.

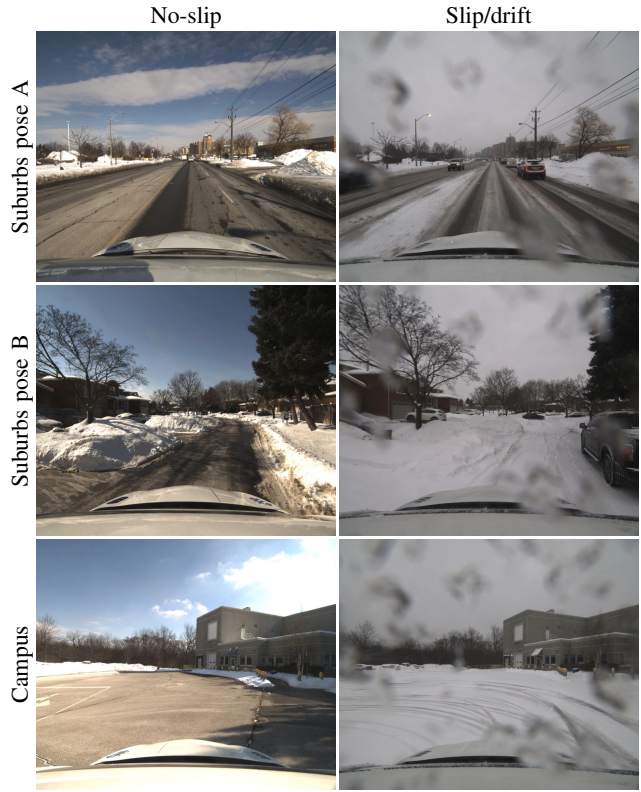
\begin{figure}
    \centering
    \def\widthfactor{0.46}
    \def\nodedist{0.05cm}
    \def\legendtextsize{\small}
    \def\legenddist{0.05cm}
    \begin{tikzpicture}
     \node[inner sep=0, outer sep=0](sAn){\includegraphics[width=\widthfactor\columnwidth]{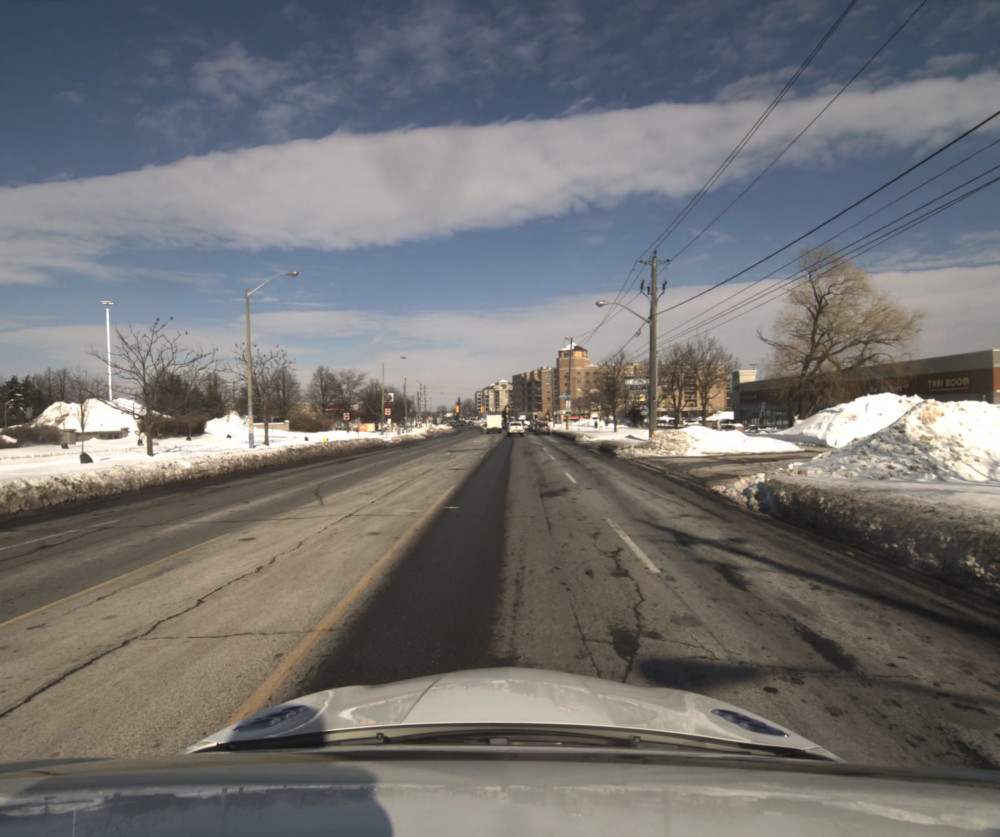}};
     \node[inner sep=0, outer sep=0, right=\nodedist of sAn](sAs){\includegraphics[width=\widthfactor\columnwidth]{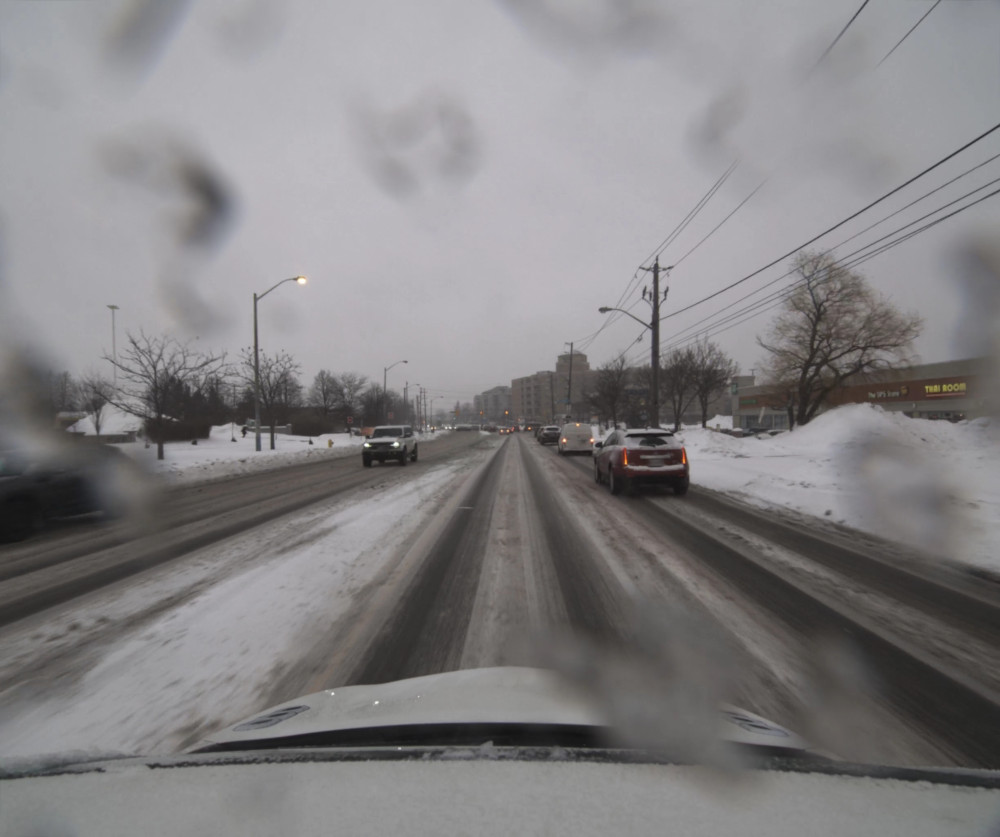}};
     \node[inner sep=0, outer sep=0, below=\nodedist of sAn](sBn){\includegraphics[width=\widthfactor\columnwidth]{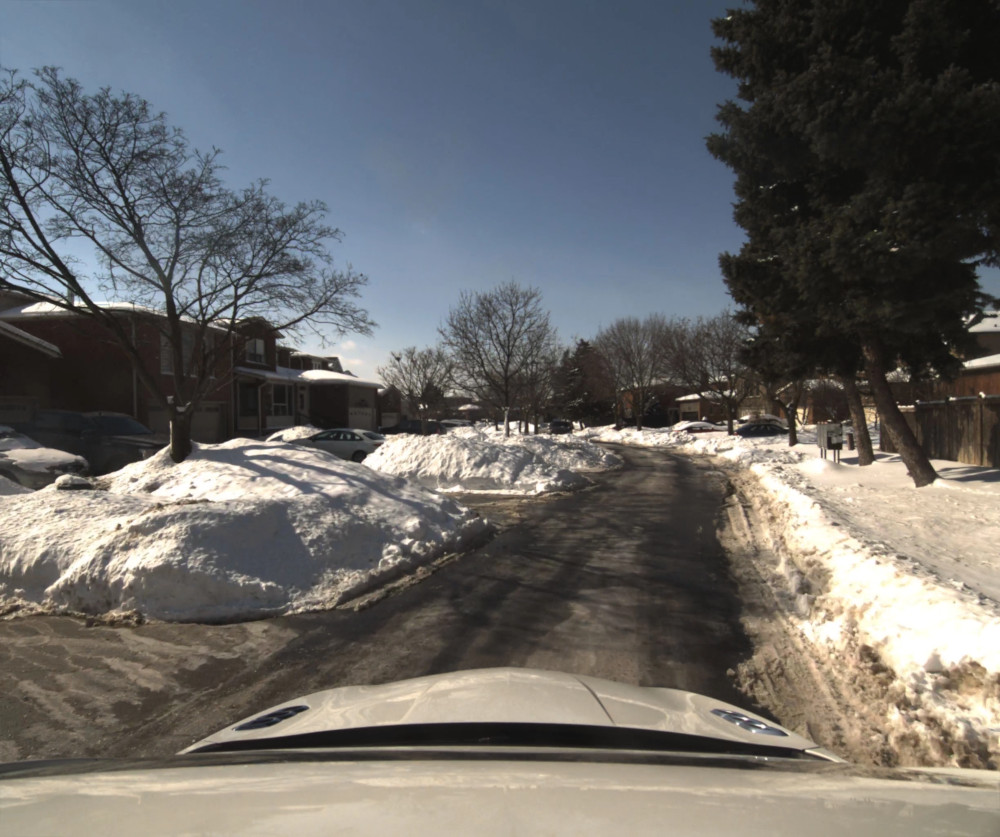}};
     \node[inner sep=0, outer sep=0, right=\nodedist of sBn](sBs){\includegraphics[width=\widthfactor\columnwidth]{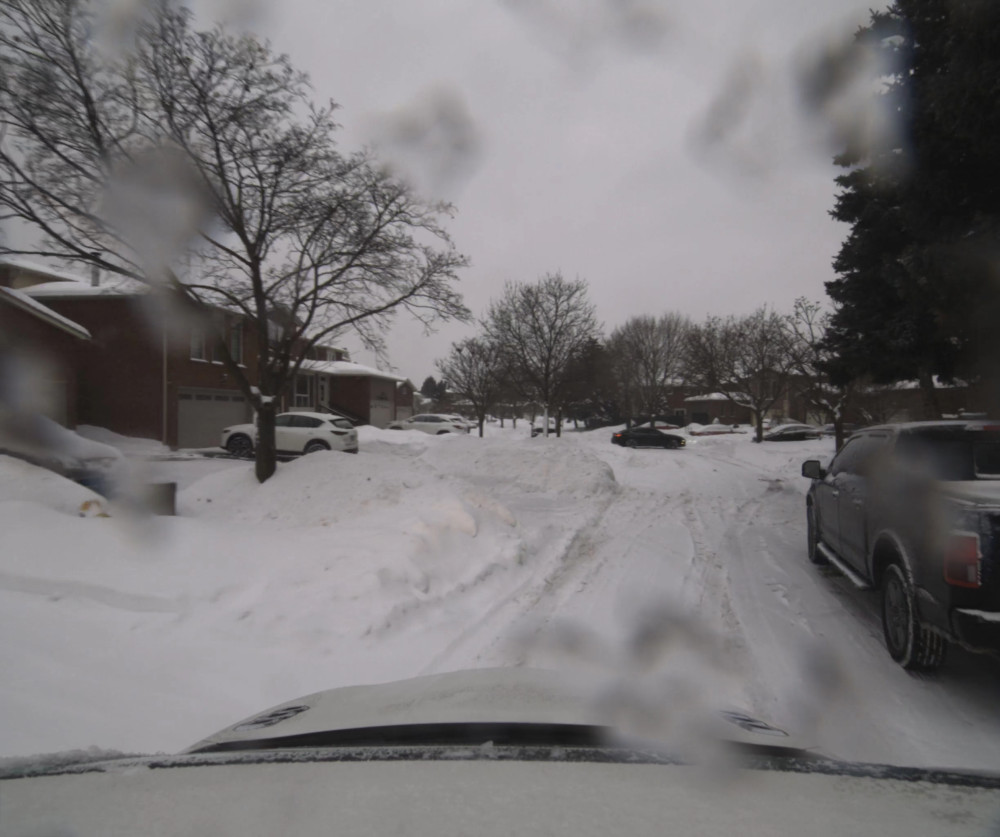}};
     \node[inner sep=0, outer sep=0, below=\nodedist of sBn](cn){\includegraphics[width=\widthfactor\columnwidth]{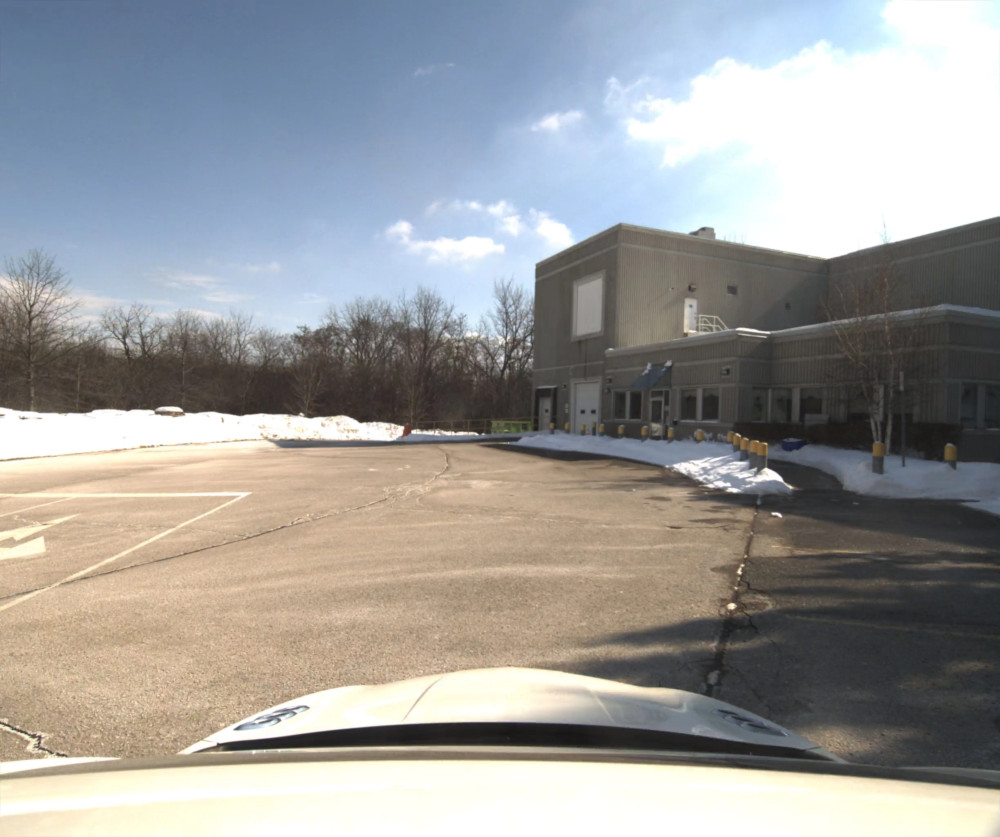}};
     \node[inner sep=0, outer sep=0, right=\nodedist of cn](cs){\includegraphics[width=\widthfactor\columnwidth]{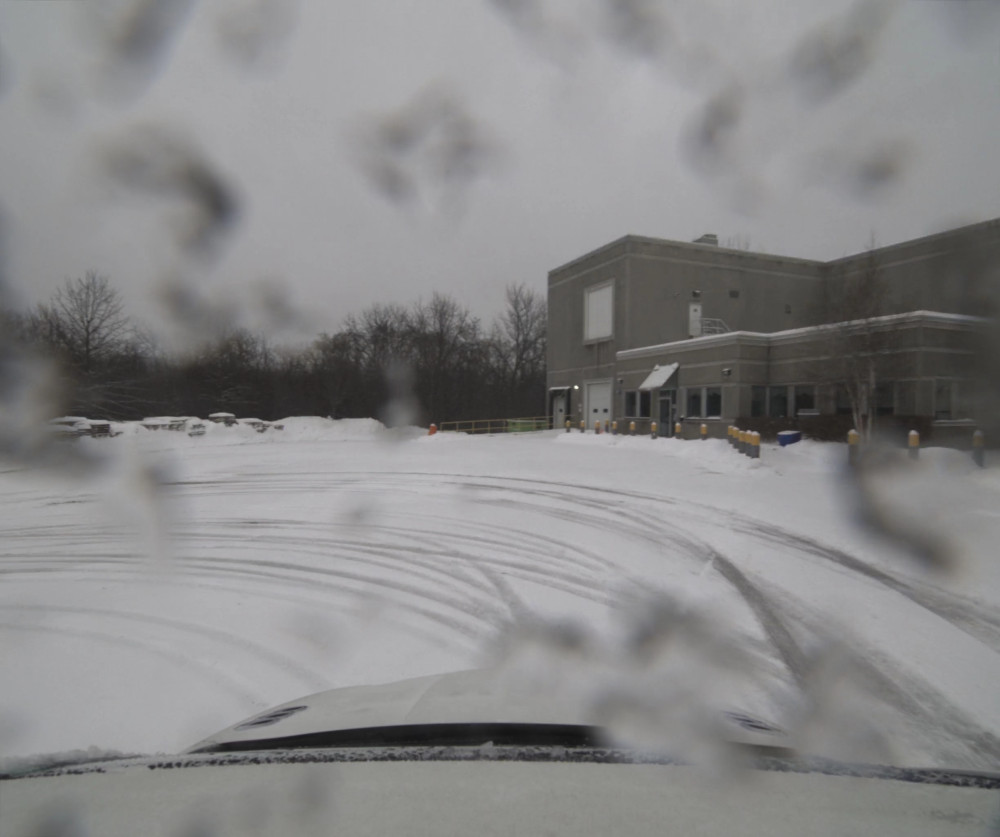}};

     \node[inner sep=0, outer sep=0, above=\legenddist of sAn]{\legendtextsize No-slip};
     \node[inner sep=0, outer sep=0, above=\legenddist of sAs]{\legendtextsize Slip/drift};
     \node[inner sep=0, outer sep=0, rotate=90, left=\legenddist of sAn, anchor=south]{\legendtextsize Suburbs pose A};
     \node[inner sep=0, outer sep=0, rotate=90, left=\legenddist of sBn, anchor=south]{\legendtextsize Suburbs pose B};
     \node[inner sep=0, outer sep=0, rotate=90, left=\legenddist of cn, anchor=south]{\legendtextsize Campus};
     
    \end{tikzpicture}
    \caption{Sample images from the front-facing camera during collection of the data used in our analysis on the impact of slippage on wheel-based odometry.}
    \label{fig:data_images}
\end{figure}

\begin{figure}
    \centering
    \includegraphics[width=0.90\columnwidth]{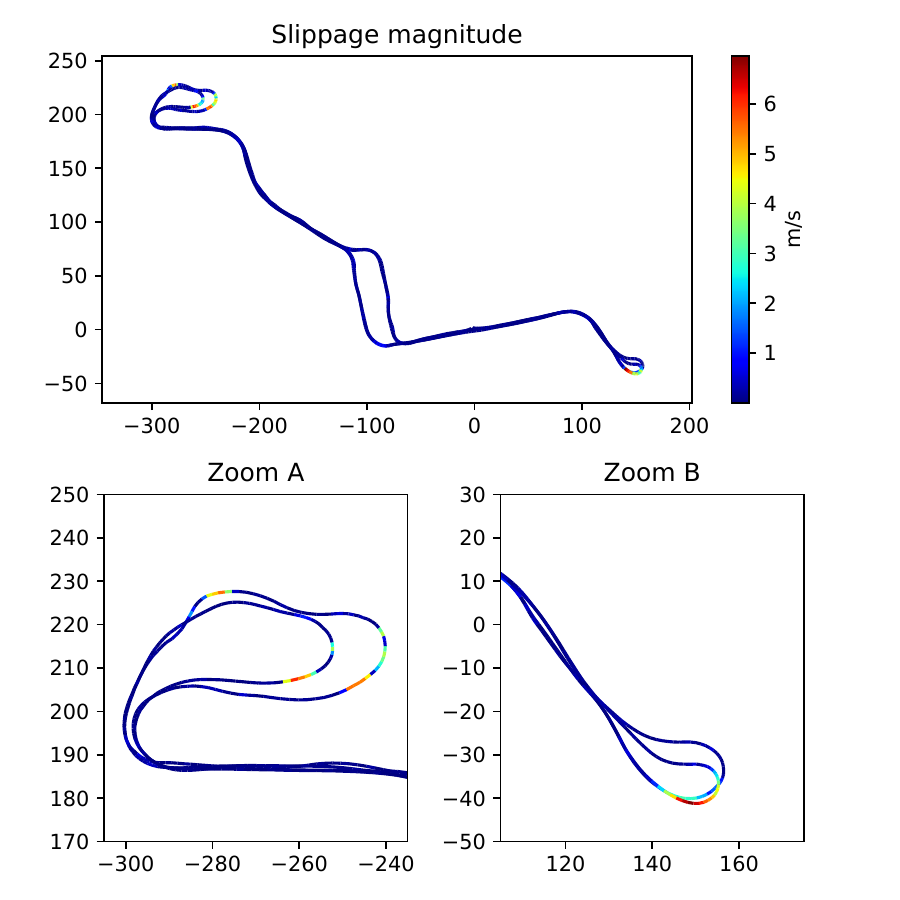}
    \vspace{-0.3cm}
    \caption{Visualization of the slippage in one of the \emph{Campus drift} sequences. The slippage is defined as the difference between the forward velocity measured by the wheel encoder and the ground-truth velocity from the GNSS/INS projected in the wheel reference frame.}
    \label{fig:campus_slippage_magnitude}
\end{figure}

In total, across February 2025, we have collected 15 new sequences as follows:
\begin{itemize}
    \item \textbf{Suburbs no-slip:} Following the Boreas route within a suburban area of Toronto, we collected three sequences in good road conditions (not driving on snow). Each sequence represents around $7.9\,\si{km}$ of driving over 15 minutes.
    \item \textbf{Suburbs slip:} We recorded two sequences on the same route as \emph{Suburbs no-slip} during a snowstorm.
    As illustrated in Fig.~\ref{fig:data_images}, the road conditions are quite challenging, with part of the road partly cleared and others covered with a thick layer of snow.
    The first sequence does not possess any voluntary or noticeable slippage, while the second one has induced slippage around the start by pulling the handbrake in a curve.
    \item \textbf{Campus no-slip:}
    Five sequences have been recorded on the University of Toronto Aerospace Institute campus by repeating a loop between two parking lots twice (cf. Fig.~\ref{fig:teaser}(b)). Each sequence lasts around 7 minutes and spans $2.7\,\si{km}$.
    \item \textbf{Campus drift:}
    These five sequences repeat the above campus loops during a snowstorm, as illustrated in Fig.~\ref{fig:data_images}.
    The vehicle is driven to purposely slip by pulling the handbrake in sharp turns and trying to maintain the slippage over increasing periods of time (slippage magnitude of up to $7\,\si{m/s}$ as shown in Fig.~\ref{fig:campus_slippage_magnitude}).
\end{itemize}
Please note that the Boreas vehicle is an \ac{awd} car.
Accordingly, we expect this platform to display higher levels of slippage than \ac{fwd} cars as engine torque is applied to the encoder-equipped wheel.
Without side-slip, a \ac{fwd} car would experience slippage only if the brake locked the encoder wheel.

\subsubsection{Baselines}

To benchmark the \ac{og} odometry, we used two radar-based baseline methods.
The first baseline denoted Radar-DG, corresponds to the best-performing method from~\cite{lisus2025doppler}.
Its core consists of an ICP-based scan registration using point clouds extracted from the raw radar data and a \ac{gp}-based continuous-time state representation.
Gyroscope measurements are preintegrated and added as extra constraints on the orientation.
The method also leverages Doppler-based velocity measurements.
The second baseline is DRO-G.
We also provide a lidar-based baseline Lidar-G by adding gyroscope constraints to the continuous-time ICP from~\cite{are_we_ready_for}.

As presented in Section~\ref{sec:og}, the \ac{og} odometry requires a calibration step.
In this setup, we use two different sets of data to estimate the wheel radius and the extrinsic parameters required for evaluation.
The first one is based on the same 14 sequences from the public Boreas dataset~\cite{burnett2023boreas} collected between December 2020 and November 2021 (thus, using the same parameters as in Section~\ref{sec:boreas_exp}).
The second calibration leverages 6 sequences collected in December 2024 in various environments similar to the ones in~\cite{lisus2025doppler} (suburban, highway, tunnel, and skyway).
We denote the \ac{og} odometry results obtained with the two calibration sets as \ac{og}$_{21}$ and \ac{og}$_{24}$, respectively.

\subsubsection{Results}

\begin{table*}[]
    \centering
    \caption{Average relative pose accuracy of the proposed method and several baselines on our dataset (\textbf{best}, \underline{second best}).}
    \vspace{-0.3cm}
    \setlength{\tabcolsep}{2pt}
    \begin{tabularx}{\linewidth}{lYYYYY}
        \toprule
        \textbf{Sequence type} \scriptsize(length, avg. / max. vel.) & \textbf{Radar-DG} \cite{lisus2025doppler} & \textbf{DRO-G} & \textbf{OG$_{21}$} & \textbf{OG$_{24}$} & \textbf{Lidar-G} \cite{are_we_ready_for}
        \\
        \midrule
        Suburbs no-slip \scriptsize($3\times7.9\, \si{\km}$, $6.9$ / $16.3\,\si{\m/\s}$)  & 0.32 / 0.04 & \underline{0.18} / \textbf{0.02} & 0.30 / \underline{0.03} & 0.24 / \underline{0.03} & \textbf{0.15} / 0.05
        \\ 
        Suburbs slip \scriptsize($2\times7.9\, \si{\km}$, $6.8$ / $16.3\,\si{\m/\s}$)  & 0.30 / \underline{0.04} & \underline{0.24} / \textbf{0.03} & 0.44 / 0.07 & 0.50 / 0.07 & \textbf{0.12} / \underline{0.04}
        \\ 
        Campus no-slip \scriptsize($5\times2.7\, \si{\km}$, $6.4$ / $12.2\,\si{\m/\s}$)  & 0.21 / \underline{0.04} & \underline{0.17} / 0.05 & 0.20 / 0.06 & 0.18 / 0.06 & \textbf{0.11} / \textbf{0.03}
        \\ 
        Campus drift \scriptsize($5\times2.7\, \si{\km}$, $6.2$ / $10.9\,\si{\m/\s}$)  & 0.21 / 0.05 & \underline{0.16} / \underline{0.04} & 1.00 / 0.07 & 1.03 / 0.07 & \textbf{0.13} / \textbf{0.02}
        \\ 
        \bottomrule
        \multicolumn{6}{c}{\scriptsize KITTI odometry metric reported as \textit{XX / YY} with \textit{XX} [\%] and \textit{YY} [$\si{\degree}/100\,\si{\m}$] the translation and orientation errors, respectively.}
    \end{tabularx}
    \label{tab:boreas}
\end{table*}

\begin{figure*}
    \centering
    \def\figscale{0.45}
    \def\hdist{0.5cm}
    \def\legendtextsize{\small}
    \def\legenddist{0.0cm}
    \begin{tikzpicture}
        \node[inner sep=0, outer sep=0](noslip){\includegraphics[width=\figscale\linewidth]{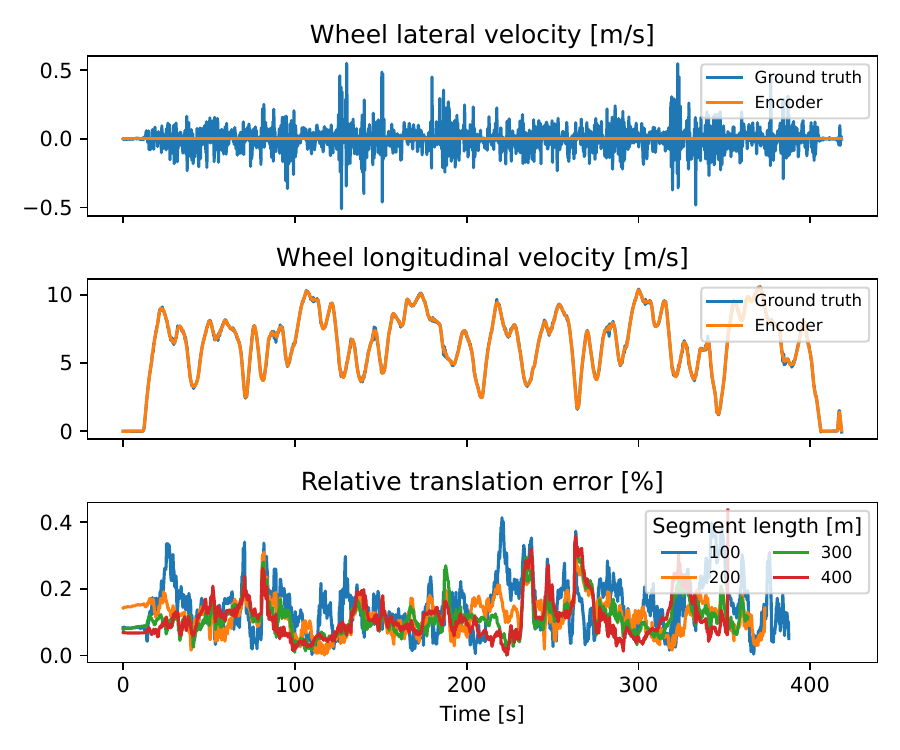}};
        \node[inner sep=0, outer sep=0, right=\hdist of noslip](drift){\includegraphics[width=\figscale\linewidth]{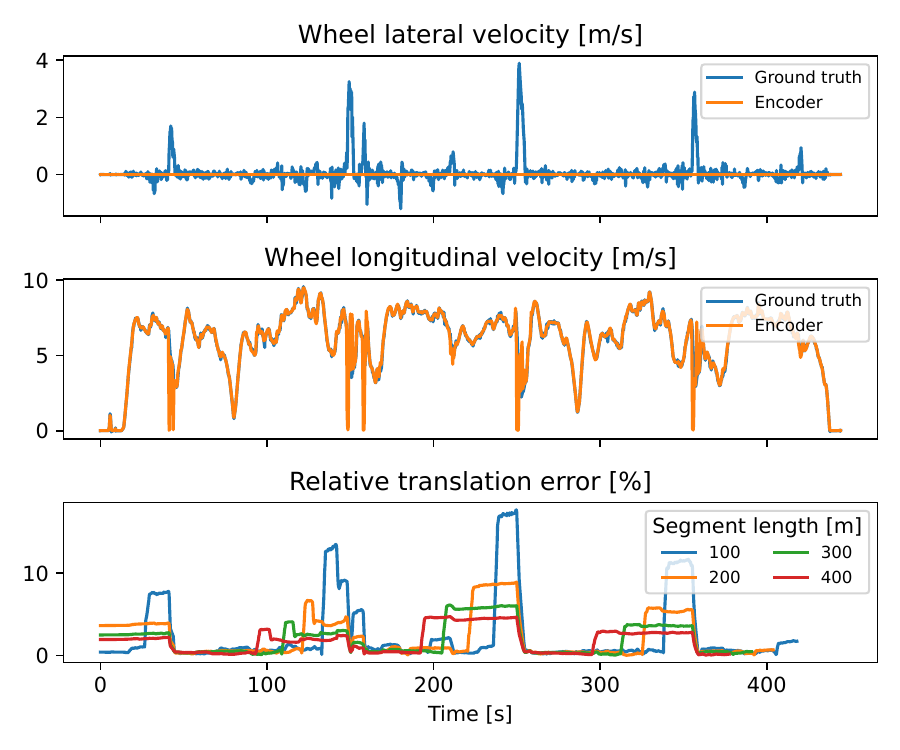}};
        \node[inner sep=0, outer sep=0, below=\legenddist of noslip]{\legendtextsize \emph{Campus no-slip 1}};
        \node[inner sep=0, outer sep=0, below=\legenddist of drift]{\legendtextsize \emph{Campus drift 4}};
    \end{tikzpicture}
    \vspace{-0.3cm}
    \caption{Visualization of the velocity measured by the wheel encoder and the ground-truth for two \emph{Campus} sequences. The last row of each plot shows the KITTI relative translation error for different segment lengths (lengths above $400\si{m}$ have been omitted for the sake of readability). One can easily see the impact of sudden handbrake-induced slippage in the velocity disparity and error jumps.}
    \label{fig:lateral_vel}
\end{figure*}

Table~\ref{tab:boreas} shows the average KITTI metrics obtained for each sequence type.
Regardless of the calibration set, the \ac{og} odometry provides accuracy on par with radar and lidar \ac{sota} methods in no-slip scenarios.
As expected, when the vehicle trajectory includes slippage, the performance of the \ac{og} odometry deteriorates while the exteroceptive methods maintain similar error levels.
The car slippage of the \emph{Campus drift} sequence shown in Fig.~\ref{fig:campus_slippage_magnitude}) has a strong, local impact on the odometry estimates, as illustrated in Fig.~\ref{fig:teaser}(c), and~\ref{fig:lateral_vel}.

An interesting observation is the weak correlation between the time of recording of the calibration sets and the overall accuracy of the \ac{og} odometry (\ac{og}$_{21}$ vs. \ac{og}$_{24}$).
There is only a small drop in performance in nominal conditions when using 3-year-old calibration parameters.
This enables the use of wheel-based odometry at the industrial scale without the burden of regular calibration.
Additionally, we believe that the calibration procedure can be easily modified to be performed continuously using noisy measurements from GPS units present in all modern vehicles, without requiring a high-precision RTK-GNSS/INS solution.
Such an approach can also enable continuous gyroscope bias estimation without requiring the vehicle to stop. 

\begin{figure}
    \centering
    \begin{tikzpicture}
        \node[inner sep = 0, outer sep = 0]{\includegraphics[width=0.99\columnwidth]{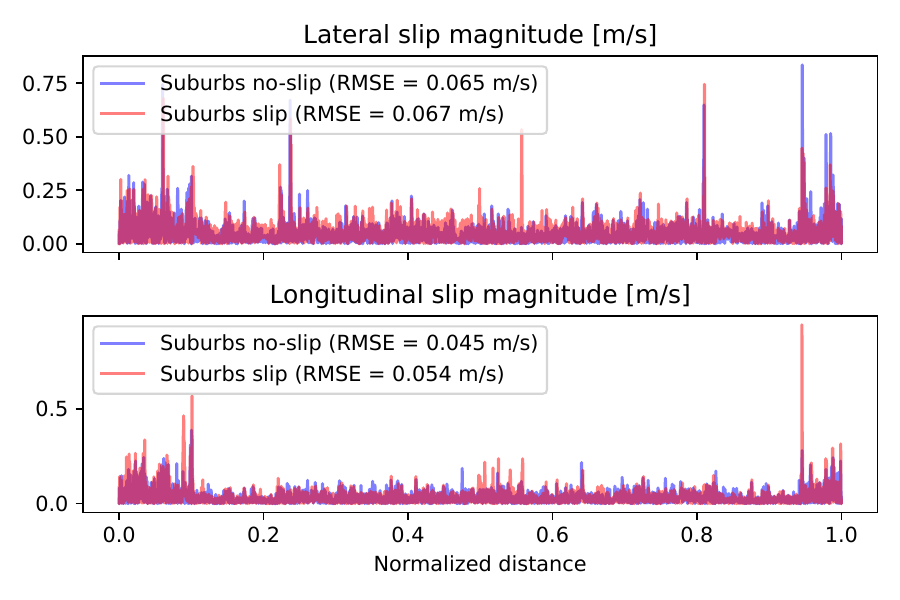}};
        \draw[decorate,decoration = {calligraphic brace}] (0.8, -2.7) --  (0.2, -2.7);
        \draw[decorate,decoration = {calligraphic brace}] (3.0, -2.7) --  (1.0, -2.7);
        \draw[decorate,decoration = {calligraphic brace}] (3.8, -2.7) --  (3.2, -2.7);
        \draw[decorate,decoration = {calligraphic brace}] (0.0, -2.7) --  (-2.2,-2.7);
        \draw[decorate,decoration = {calligraphic brace}] (-2.4,-2.7) --  (-3.2,-2.7);
        \node[inner sep=0, outer sep=0] at (0.5, -3.0) {\scriptsize Thick snow};
        \node[inner sep=0, outer sep=0] at (-2.8,-3.0) {\scriptsize Thick snow};
        \node[inner sep=0, outer sep=0] at (3.5, -3.0) {\scriptsize Thick snow};
        \node[inner sep=0, outer sep=0] at (2.0, -3.0) {\scriptsize Slush};
        \node[inner sep=0, outer sep=0] at (-1.1,-3.0) {\scriptsize Slush};
    \end{tikzpicture}
    \caption{Magnitude of lateral and longitudinal slippage for the best \emph{Suburbs no-slip} and \emph{Suburbs slip} sequences (with the slippage defined as the difference between the body-centric velocities measured with the encoder and the GNSS/INS solution). The bottom row describes the road conditions for the different segments of the route.}
    \label{fig:suburbs_slippage}
\end{figure}

While the \emph{Campus drift} sequences are not representative of normal driving behaviours, the \emph{Suburbs slip} ones are.
Table~\ref{tab:slippage} shows the slippage characteristics and odometry results for the independent slip and drift sequences.
One needs to keep in mind that there is a certain level of noise on the ground-truth velocity, and the current analysis does not account for the roll of the car's body.
Thus, the lateral slip is not exactly zero in no-slip scenarios.
As expected, the error of the \ac{og} odometry on the \emph{Campus drift} sequences increases with the \ac{rms} slippage.
Interestingly, the results obtained with the \emph{Suburbs slip} sequences are similar despite the intentional slip at the beginning of \emph{Suburbs slip 2}.
While the drop in performance compared to the non-slip data is significant, a relative translation error of 0.50\% stays competitive with respect to the \ac{sota} radar methods on the Boreas leaderboard.
We further the analysis by displaying the slippage magnitude of the first slip and no-slip sequences in Fig.\ref{fig:suburbs_slippage}.
One can see that the lateral-slip magnitude is similar for both sequences with an RMSE difference of only $2\,\si{mm/s}$, whereas the longitudinal slippage differs by almost $1\,\si{cm/s}$.
Relative to the average velocity of $6.8\,\si{m/s}$, it represents a forward velocity error of around 0.14\%, which is in the same order of magnitude as the difference in the odometry error.
Another point highlighted by Fig.~\ref{fig:campus_slippage_magnitude} is that the longitudinal-slip magnitude mainly differs in specific route segments with different road conditions.
The first row of Fig.~\ref{fig:data_images} illustrates the `slush' conditions, while the second row is an example of `thick snow'.
Accordingly, given normal driving behaviour, only a thick snow layer on the ground significantly affects the odometry performance.

\begin{table*}[]
    \centering
    \caption{Slippage characteristics and odometry results.}
    \vspace{-0.3cm}
    \setlength{\tabcolsep}{2pt}
    \begin{tabularx}{\linewidth}{lYYYYYY}
        \toprule
        \textbf{Sequence} & \textbf{RMS side slip} & \textbf{RMS forward slip} &\textbf{Radar-DG} \cite{lisus2025doppler} & \textbf{DRO-G} & \textbf{OG$_{24}$} & \textbf{Lidar-G} \cite{are_we_ready_for}
        \\
        \midrule
        Suburbs slip 1 & 0.067 & 0.054 & 0.27 / 0.04 & 0.18 / 0.02 & 0.50 / 0.09 & 0.11 / 0.04
        \\
        Suburbs slip 2 & 0.179 & 0.084 & 0.32 / 0.04 & 0.30 / 0.05 & 0.50 / 0.05 & 0.13 / 0.04
        \\
        \midrule
        Suburbs no-slip avg. & 0.060 & 0.047 & 0.32 / 0.04 & 0.18 / 0.02 & 0.24 / 0.03  & 0.15 / 0.05
        \\ 
        \midrule
        Campus drift 1 & 0.108 & 0.104 & 0.21 / 0.04 & 0.17 / 0.03 & 0.28 / 0.05 & 0.11 / 0.02
        \\
        Campus drift 2 & 0.222 & 0.173 & 0.17 / 0.03 & 0.16 / 0.04 & 0.50 / 0.05 & 0.11 / 0.02
        \\
        Campus drift 3 & 0.422 & 0.218 & 0.19 / 0.04 & 0.17 / 0.05 & 1.09 / 0.15 & 0.13 / 0.02
        \\
        Campus drift 4 & 0.652 & 0.437 & 0.22 / 0.05 & 0.17 / 0.04 & 1.95 / 0.04 & 0.16 / 0.02
        \\
        Campus drift 5 & 0.520 & 0.325 & 0.27 / 0.08 & 0.15 / 0.07 & 1.32 / 0.06 & 0.14 / 0.02
        \\
        \midrule
        Campus no-slip avg. & 0.101 & 0.095 & 0.21 / 0.04 & 0.17 / 0.05 & 0.18 / 0.06 & 0.11 / 0.03
        \\ 
        \bottomrule
        Units &  [$\si{m/s}$] &  [$\si{m/s}$] & \multicolumn{4}{c}{KITTI odometry metric reported as \textit{XX / YY} with \textit{XX} [\%] and \textit{YY} [$\si{\degree}/100\,\si{\m}$]}
    \end{tabularx}
    \label{tab:slippage}
\end{table*}

A very important difference between the methods benchmarked in this section is the computational load.
Table~\ref{tab:computation_time} reports the `per-frame' computation time of each of the methods (our implementation outputs the \ac{og} estimates at the frequency of the radar for ease of evaluation).
DRO uses an Nvidia RTX 5000 (Mobile) GPU while all the other methods run on a single core of an Intel i7-12800H CPU.
Our naive Python implementation of the \ac{og} odometry displays a negligible execution time compared to any of the baselines.
Additionally, as the typical sampling rate of gyroscopes and encoders is higher than that of exteroceptive sensors, the \ac{og} estimates can be provided at high frequency, leading to lag-less pose information.

\begin{table}[]
    \centering
    \caption{Per-frame computation time}
    \vspace{-0.3cm}
    \begin{tabularx}{\linewidth}{lYYYY}
        \cmidrule[\heavyrulewidth]{2-5}
         &\textbf{Radar-DG} \cite{lisus2025doppler} & \textbf{DRO-G} & \textbf{OG} & \textbf{Lidar-G} \cite{are_we_ready_for}
         \\
         \midrule
         \textbf{Data span} [$\si{ms}$] & 250 & 250 & 250 & 100
         \\
         \textbf{Exec. time} [$\si{ms}$] & 110 &  115 & $<$1 & 201
         \\
         \bottomrule
    \end{tabularx}
    \label{tab:computation_time}
\end{table}

\section{Discussion}
\label{sec:discussion}

The results obtained in our experiments lead us to a simple question: Do we still need to work on odometry for autonomous driving applications?
After more than a decade of research, hardware improvement, and processing power increase, we witness a diminishing return on accuracy gain vs. resources spent on developing newer odometry frameworks for automotive applications.
At first, we were very surprised by our findings.
Despite all the past efforts, a simple encoder-gyroscope approach is on par with \ac{sota} exteroceptive odometry.
It is a reminder that one should try the easy solutions first before investing time and money into high-complexity systems.
Except for very niche scenarios such as thick snow cover, where the \ac{og} odometry can be challenged, there is no rational explanation for why we should use much more complex, expensive, and sometimes less reliable solutions.
Accordingly, our first takeaway is that the automotive robotics community should shift away from the problem of odometry in nominal scenarios where simple solutions work well enough.
Instead, we should focus on edge cases where simple methods are more limited (such as the problem of slip detection for the \ac{og} odometry).
We call for the publication of quality datasets and benchmarks that expose such scenarios.
Please note that the general problem of odometry is still quite relevant in other applications where wheel-based odometry is not possible (e.g., UAVs and handheld devices) or when the amount of slippage is very high (e.g., off-road robots).

On top of the aforementioned considerations, we believe that the problem of odometry and using its accuracy as a measure of performance and readiness for real-world deployment is missing part of the big picture.
While odometry and \ac{slam} accuracy are crucial for autonomous exploration\footnote{One can think about autonomous vacuum cleaners that map the environment before switching to an `exploitation' phase.}, it is not necessarily the case for autonomous driving, where the platform is expected to navigate between end goals set in a prior map.
As mapping the environment can be done with specialized platforms that include high-accuracy GNSS/INS solutions, solving the problem of large-scale odometry (ego-motion estimation in an unknown environment) is not the most relevant problem.
The robust localization within a prior map and the ability to update such a map when changes occur in the environment are key to autonomous driving.
In such a context, odometry is `only' needed locally to ensure temporal consistency of the pose estimates and improve computational efficiency with better initial guesses and keyframing.
Despite the high relevance of the topic, localization is generally less researched and only a few works, such as~\cite{are_we_ready_for}, address it in the context of self-driving vehicles.

\section{Conclusion}
\label{sec:conclusion}

In the context of autonomous driving, we have explored the possibility of performing highly accurate odometry using solely a wheel encoder and a yaw gyroscope (OG odometry).
Given nominal vehicle behaviour (limited wheel slippage), the OG odometry outperforms some of the exteroceptive baselines for a tiny fraction of the computational cost.
Legitimately, one can wonder about the added value of further research on the topic.
Attempting to answer this question, we have pushed the limits of wheel-based odometry by intentionally violating the main assumption with the introduction of slippage in the vehicle trajectory.
In such scenarios, the performance of the OG odometry drops significantly, while other methods keep performing at satisfying levels of accuracy.

This work highlights the need for datasets with off-nominal driving behaviours, and we advocate shifting away from general odometry research for autonomous driving and moving towards more relevant and specific challenges.
The first challenge is vehicle-slip detection and accounting for it within a wheel-based odometry framework in a cost- and computation-efficient manner.
This would not only be a step forward for autonomous navigation, but it would also be beneficial for \acp{adas}.
Another relevant problem is global and/or multi-session localization.
To enable real applications, autonomous vehicles need to navigate between positions known within a prior map.
The odometry estimation in `unknown environments' alone does not allow for the execution of such high-level tasks.

Accordingly, our future work includes collecting and releasing a high-quality dataset containing different off-nominal trajectories in challenging environments, including the sequences used in this paper.
We will explore the performance of the OG odometry in 3D using a three-axis gyroscope before deriving a slip-detection algorithm by adding a three-axis accelerometer.
Finally, on the localization topic, we will integrate the \ac{og} odometry within a \emph{teach and repeat} system for highly efficient multi-session localization.



\bibliographystyle{IEEEtran}
\bibliography{references}

\end{document}